\documentclass[conference]{IEEEtran}
\IEEEoverridecommandlockouts
\usepackage{cite}
\usepackage{amsmath,amssymb,amsfonts}
\usepackage{algorithmic}
\usepackage{graphicx}
\usepackage{textcomp}
\usepackage{xcolor}
\usepackage{url}
\usepackage{multirow}
\usepackage{lscape}
\usepackage[caption=false]{subfig}
\begin{document}

\title{Solving the QAP by Two-Stage Graph Pointer Networks and Reinforcement Learning}
\author{
\IEEEauthorblockN{Satoko Iida and Ryota Yasudo}
\IEEEauthorblockA{Kyoto University, Japan\\Email: yasudo@i.kyoto-u.ac.jp}
}

\maketitle

\begin{abstract}
Quadratic Assignment Problem (QAP) is a practical combinatorial optimization problems that has been studied for several years.
Since it is NP-hard, solving large problem instances of QAP is challenging. Although heuristics can find semi-optimal solutions, the execution time significantly increases as the problem size increases.
Recently, solving combinatorial optimization problems by deep learning has been attracting attention as a faster solver than heuristics. Even with deep learning, however, solving large QAP is still challenging.
In this paper, we propose the deep reinforcement learning model called the two-stage graph pointer network (GPN) for solving QAP.
Two-stage GPN relies on GPN, which has been proposed for Euclidean Traveling Salesman Problem (TSP).
First, we extend GPN for general TSP, and then we add new algorithms to that model for solving QAP.
Our experimental results show that our two-stage GPN provides semi-optimal solutions for benchmark problem instances from TSPlib and QAPLIB.
\end{abstract}

\section{Introduction}
Combinatorial optimization is a fundamental problems in computer science and has been studied for long years.
Traditionally, the methods to solve NP-hard problems are divided into two classes, {\em exact} and {\em heuristic} methods. The exact methods take the time to find the optimal solutions with algorithmic techniques such as branch and bound. While the solutions given by such methods are guaranteed to be exactly optimal, it is not practical for large problems because of the long execution time. On the contrary, the heuristic methods attempt to find semi-optimal solutions quickly with metaheuristics such as the local search~\cite{2-opt, LKH}. Although the execution time is shorter than that of exact methods, the execution time significantly increases as the size of problems increases.

Recently, machine learning has been gathering attention as the third method to solve combinatorial optimization problems as well as finding novel algorithms~\cite{fawzi2022,mankowitz2023} and a chip design~\cite{mirhoseini2021}. This method, called the neural combinatorial optimization, makes machine learning learn parameters of a neural network so that it outputs a semi-optimal solution. The advantages of this method include the speed and the scalability. First, once the training has been finished, the inference that generates the solution requires little execution time. This is much shorter than that of heuristic methods. Secondly, the execution time increases only slightly even if the size of problems increases. In particular, recurrent neural networks enable us to solve problems with arbitrary number of variables with one neural network model trained with small problems.

In particular, the pointer network (Ptr-Net) has been proposed. It learns the conditional probability of an output sequence with elements, and hence they can be used for solving traveling salesman problem (TSP) as well as convex hull identification~\cite{ptr}. Subsequently the graph pointer network (GPN) adds graph neural networks (GNN) as a graph embedding layer to Ptr-Net and improves the solution quality~\cite{GPN}. However, both Ptr-Net and GPN solve only 2D Euclidean TSP, where the input is XY cordinates of each city and the distance between two cities is equal to the Euclidean distance. Thus, the two methods cannot solve general TSP, where the instance is defined by an $n \times n$ distance matrix consisting of the distances between any two cities. GPN should be generalized so that it can solve such a TSP instance.

Furthermore, we can extend the network model so that it can solve the quadratic assignment problems (QAP), because QAP can be regarded as a special case of TSP and has many applications in our real world~\cite{mangoubi1985,dell2009,hubert1976,alkaya2015,steinberg1961} . The instance of QAP is given by two matrices, an $n \times n$ distance matrix and an $n \times n$ flow matrix. The cost function of QAP is the sum of the product of the distance and the flow between two factories. TSP can be regarded as a QAP such that $n$ elements, which constitutes a cycle, in the flow matrix is one. In this context, we firstly extend GPN for matrix input TSP, and then extend it for QAP. QAP is computationally more expensive than TSP because the number of possible sum of the distance and the flow is $n^4$. To solve QAP efficiently, we propose the two-stage graph pointer network.

To sum up, the main contributions of this paper are as follows.
\begin{itemize}
\item We propose an extended GPN that solves matrix input TSP where the distances between two cities are explicitly given by a distance matrix.
\item We further extend our model so that it can solve QAP with the two-stage graph pointer network.
\item We demonstrate that the two-stage graph pointer network can approximately solve QAP with arbitrary size.
\end{itemize}

The rest of the paper is organized as follows.
Section~\ref{sec:before} introduces reinforcement learning and Graph Pointer Network.
In Section~\ref{method}, we explain the proposed network model.
We implement the proposed model on GPU and evaluation on the benchmark in Section~\ref{sec:ex}.
Finally, Section~\ref{sect:conc} concludes the paper.

\section{Traveling Salesman Problem and Quadratic Assignment Problem}\label{sec:problem}

This section describes the Euclidean TSP, the matrix input TSP, and QAP.  The matrix input TSP can be regarded as a generalization of the Euclidean TSP, and QAP is an even more generalized problem.

\subsection{Euclidean TSP}

An instance of the Euclidean TSP is given by XY coordinates $\{\boldsymbol{x}_1, \boldsymbol{x}_2, \ldots, \boldsymbol{x}_N \}$ of $N$ cities. The distance between two cities is equal to the Euclidean distance $\| \boldsymbol{x}_i - \boldsymbol{x}_j \|_2$. The goal of the Euclidean TSP is to find the permutation $\sigma$ over the cities that minimizes the tour length
\begin{eqnarray}
\sum_{i=1}^{N} \| \boldsymbol{x}_{\sigma(i)} - \boldsymbol{x}_{\sigma(i+1)}\|_2,
\end{eqnarray}
where $\sigma(1) = \sigma(N+1)$, $\sigma(i) \in \{1, 2, \ldots, N\}$, and $\sigma(i) \neq \sigma(j)$ for any $i \neq j$. In other words, the objective function is the length of Hamiltonian cycle.

\subsection{Matrix input TSP}
The matrix input TSP requires an $N \times N$ {\em distance} matrix $\begin{bmatrix}d_{i,j}\end{bmatrix}_{1 \leq i,j \leq N}$ as an input. This matrix contains the distances $d_{i,j}$ between the pair of cities, $i$ and $j$.
The goal of the matrix input TSP is to find the permutation $\sigma$ over the cities that minimizes the tour length
\begin{eqnarray}
\sum_{i=1}^{N} d_{\sigma(i), \sigma(i+1)},
\end{eqnarray}
where $\sigma(1) = \sigma(N+1), \sigma(i) \in \{1, 2, \ldots, N\}$, and $\sigma(i) \neq \sigma(j)$ for any $i \neq j$.
The matrix input TSP includes the Euclidean TSP because the Euclidean TSP can easily be converted by computing the Euclidean distances between any two cities. Conversely, there exists the matrix input TSP that cannot be converted to the Euclidean TSP, because of the triangle inequality.

\subsection{QAP}
An instance of the quadratic assignment problem (QAP) is given by two matrices. First, the distance matrix $\begin{bmatrix}d_{i,j}\end{bmatrix}_{1 \leq i, j \leq N}$ contains the distances for $N$ locations. Second, the {\em flow} matrix $\begin{bmatrix}f_{i,j}\end{bmatrix}_{1 \leq i, j \leq N}$ contains the flows for $N$ factories.
The objective of QAP is to find permutation $\pi$ that minimizes the total cost, computed by
\begin{eqnarray}
\sum_{i=1}^{n} \sum_{j=1}^{n} d_{\pi(i),\pi(j)} f_{i,j},
\label{eq-qap}
\end{eqnarray}
where $\pi(i) \in \{1,2, \ldots, N\}$ and $\pi(i) \neq \pi(j)$ for any $i \neq j$.

\begin{figure}[t]
    \centering
    \subfloat[]{
    \includegraphics[width=0.9\hsize]{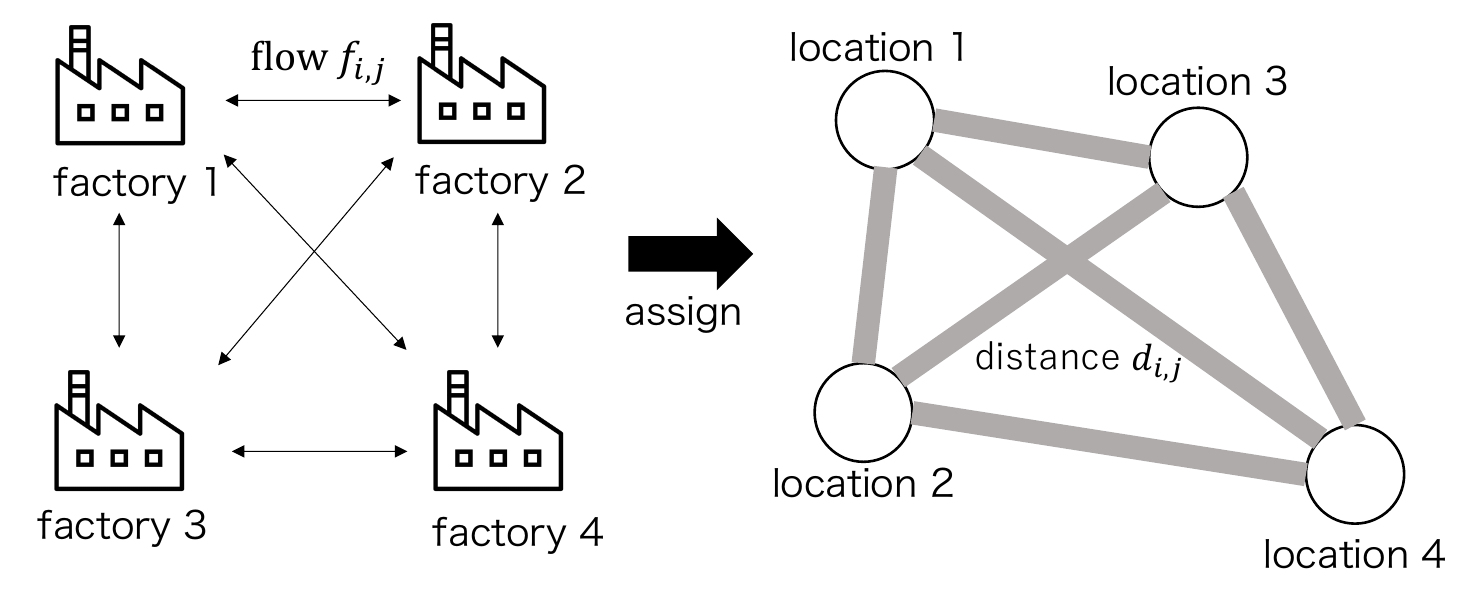}
    }\\
    \subfloat[]{
    \includegraphics[width=0.9\hsize]{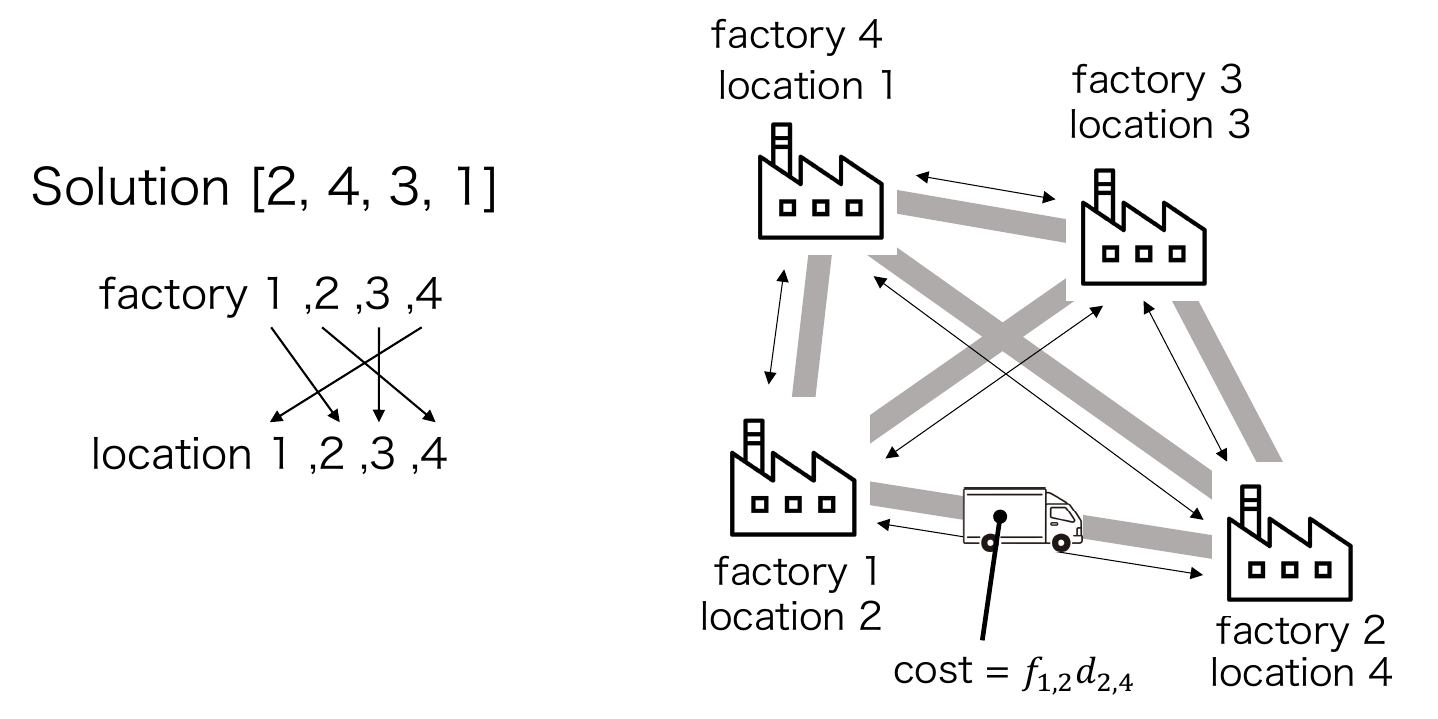}
    }
    \caption{An example of QAP. (a) Problem and (b) Solution $[2,4,3,1]$.}
    \label{fig:qap_example}
\end{figure}

We illustrate an example of QAP in Fig.~\ref{fig:qap_example}.
In this example, $f_{i,j}$ corresponds to the transportation amount between factory $i$ and factory $j$, and $d_{i,j}$ corresponds to the distance between location $i$ and location $j$.
The cost is defined as the sum of the products of the transportation amount and the distance.
The solution shown in this figure is [2,4,3,1], which means that factories 1, 2, 3, and 4 are located in locations 2, 4, 3, and 1, respectively.

It should be noted that TSP corresponds to a special case of QAP. TSP is equivalent to a QAP such that 
\begin{eqnarray}
    f_{i,j} = \begin{cases}
    1 & \text{if~} j = (i + 1) \bmod n\\
    0 & \text{otherwise}
    \end{cases}
\end{eqnarray}
and $d_{i,j}$ is a distance between two cities.
If location $i$ is assigned to factory $j$, city $i$ is visited in the $j$-th order.

\section{Conventional Pointer Networks}\label{sec:before}

\subsection{Pointer Network}
The pointer network (Ptr-Net) is a neural architecture to learn the conditional probability of an output sequence with elements~\cite{ptr}. It is based on recurrent neural networks, and the input size is variable. Thus, once a training phase is performed, an inference phase can solve combinatorial optimization problems with arbitrary size. A mechanism of neural attention makes the output correspond to positions in an input sequence. These characteristics enable Ptr-Net to solve three combinatorial optimization problems: finding planar convex hulls, computing Delaunay triangulation, and the traveling salesman problem. The model is trained by the Actor-Critic algorithm.
In~\cite{ptr}, small scale TSP with up to only 50 cities has been approximately solved by Ptr-Net.

\subsection{Graph Pointer Network}
\begin{figure}[t]
\centerline{\includegraphics[width=\hsize]{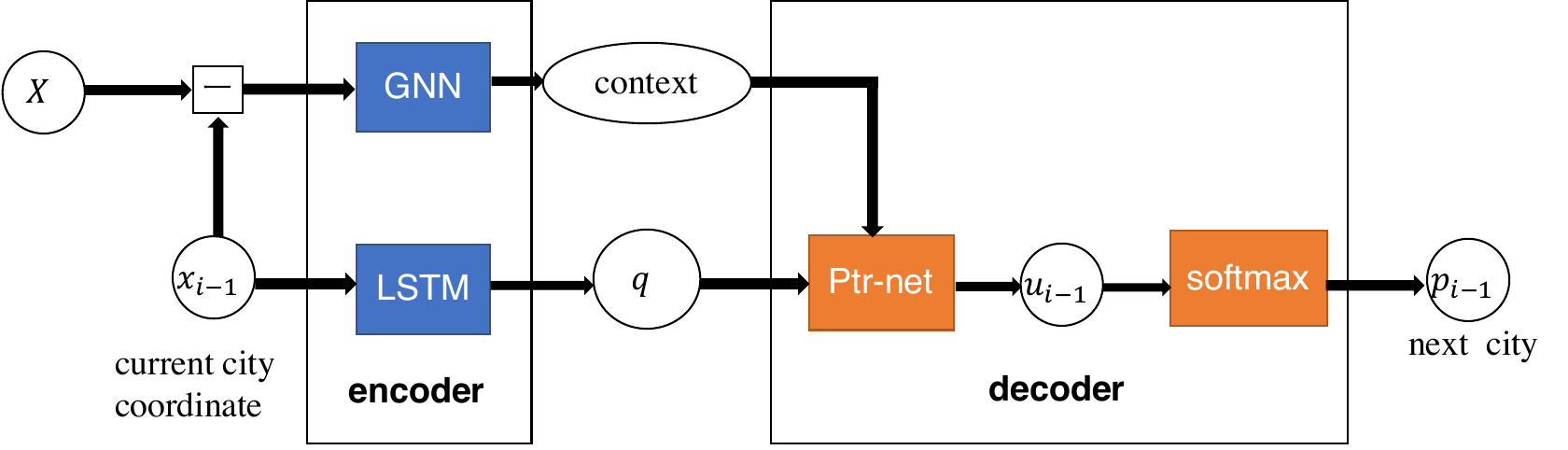}}
\caption{Overview of a graph pointer network.}
\label{fig:gpn}
\end{figure}

For improving the solution quality of Euclidean TSP, the graph pointer network (GPN) has been proposed.
GPN has an additional graph neural network layer.
GPN architecture consists of an encoder and decoder component as shown in Fig.~\ref{fig:gpn}.

The encoder consists of a Long Short-Term Memory (LSTM) and a Graph Neural Network (GNN).
The encoder receives the current city coordinates and the global city coordinates.
The current city coordinates are input to the LSTM, and the global city coordinates minus the current city coordinates are input to the GNN.
Each layer of the GNN is represented by
\begin{eqnarray}
\boldsymbol{x}_{i}^{l} = \gamma \boldsymbol{x}_{i}^{l-1} \Theta + (1 - \gamma) \phi_{\theta} \Bigl(\frac{1}{|\mathcal{N}(i)|}\{\boldsymbol{x}_{j}^{l-1}\}_{j \in \mathcal{N}(i) \cup {i }} \Bigr) .
\end{eqnarray}
Here, $\boldsymbol{x}_{i}^{l} \in \mathbb{R}^{d_{i}}$ is the $l$-th layer variable in $x_{i}^{0} = x_{i}, l \in \{ 1, \dots,L\}$, $\gamma$ is a learnable parameter to normalize the eigenvalues of the weight matrix, and
$\Theta \in \mathbb{R}^{d_{l-1} \times d_{l}}$ is a learnable weight matrix, and $\textit{N}(i)$ is the adjacency set of node $i$, and $\phi_{\theta} \colon \mathbb{R}^{d_{l-1}} \rightarrow \mathbb{R}^{d_{l}}$ 
is an aggregate function represented by a neural network.

Furthermore, considering that the graph of a symmetric TSP is a complete graph, the graph embedding layer is represented by
\begin{eqnarray}
    X^{l} = \gamma X^{l-1} \Theta + (1 - \gamma) \Phi_{\theta} \Bigl(\frac{X^{l-1}}{|\mathcal{N}(i)|} \Bigr) ,
\end{eqnarray}
where $X^{l} \in \mathbb{R}^{N \times d_{l}}$ holds, and $\phi_{\theta}\colon \mathbb{R}^{N \times d_{l-1}} \rightarrow \mathbb{R}^{N \times d_{l}}$ is the aggregate function.
This means that the LSTM layer deals with the current state, while the graph neural network layer deals with the relationship with other cities.
Afterwards, context and ref are encoded respectively and passed to the decoder.

The decoder layer takes the two features of the encoder layer as input, and outputs the features by means of an attention function.
The output vector $u_i$ is defined as
\begin{eqnarray}
 u_i =\begin{cases}
v^{T} \cdot \tanh{W_rr_j + W_qq} & \text{if~}  j\ne \sigma(k), \\
-\infty    & \text{otherwise},
\end{cases}
\end{eqnarray}
where $W_r$ and $W_q$ are trainable matrices, $q$ denotes a query vector from LSTM, and $r_i$ is a reference vector containing the context.
This vector $u_i$ is then passed through the softmax layer to obtain the policy $p_i$ as follows.
\begin{eqnarray}
    \pi_{\theta}(a_i | s_i)= p_i = \text{softmax}(u_i) .
\end{eqnarray} 
This policy $\pi_{\theta}(a_i | s_i)$ determines the next visited city $a_i$, and finally we obtain a permutation solution.

The reward function is defined as the negative cost from taking action (selecting the next city).
Hence the reward is $-\| \boldsymbol{x}_{\sigma(i)} - \boldsymbol{x}_{\sigma(i+1)}\|_2$ when the next city $\sigma(i+1)$ is selected. In this way, reinforcement learning that maximizes the reward corresponds to TSP that minimizes the tour length.
\section{Methods}
\label{method}
This section introduces our proposed model.
We begin with extending the original GPN for matrix input TSP in \ref{method-TSP}, and then introduce the proposed model for QAP in \ref{method-QAP}.

\subsection{GPN for matrix input TSP}
\label{method-TSP}
 \begin{figure*}[t]
    \centering
    \includegraphics[width=\hsize]{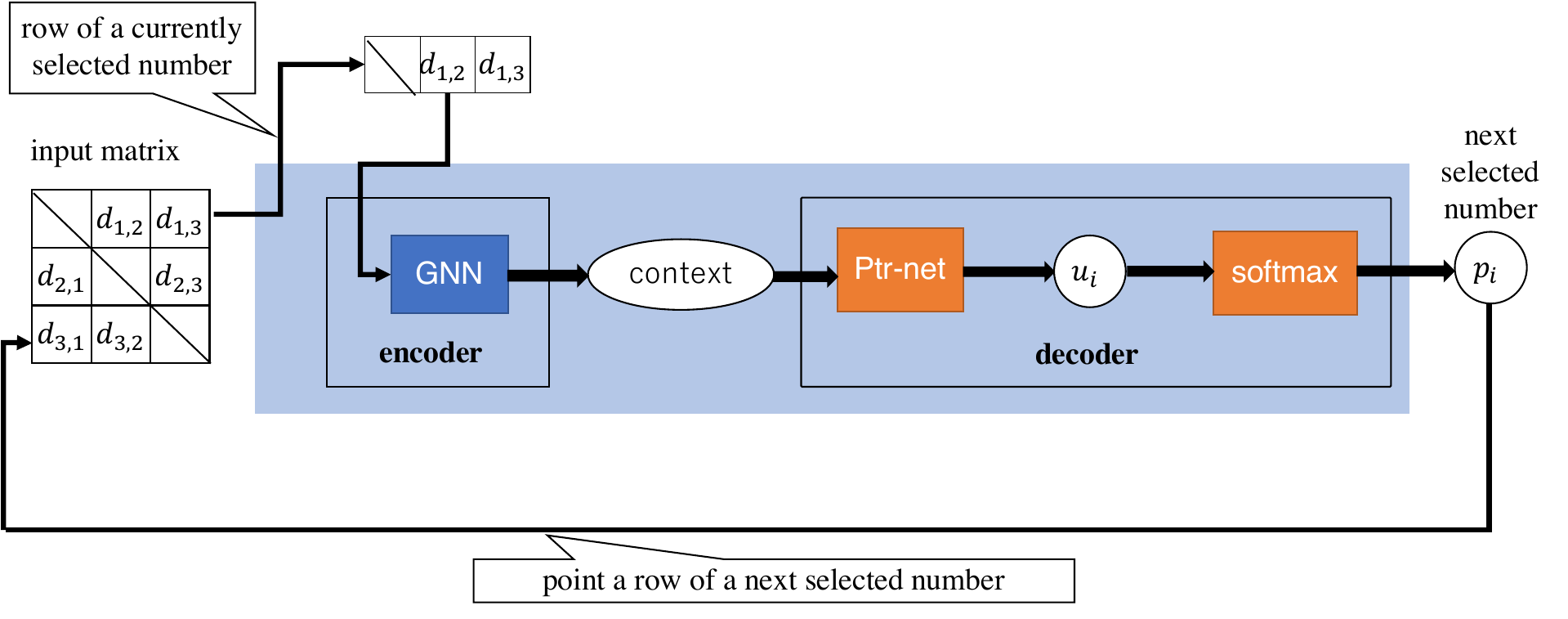}
    \caption{GPN for matrix input TSP.}
    \label{fig:mattsp}
\end{figure*}

We illustrate GPN for solving matrix input TSP in Fig.~\ref{fig:mattsp}.
It consists of an encoder and a decoder as with the original GPN, but there are several differences.
First, a row of a currently selected number from the input matrix is into the encoder.
Second, the encoder consists of only GNN: while an encoder involves LSTM in the original GPN, we propose to eliminate it.
Thus, the output vector $u_i$ is changed to
\begin{eqnarray}
 u_i =\begin{cases}
v^{T} \cdot \tanh{W_rr_j} & \text{if~}  j\ne \sigma(k), \\
-\infty    & \text{otherwise}.
\end{cases}
\end{eqnarray}

Let us now revisit the original GPN.
In the original GPN, current coordinates are entered into LSTM, and current coordinates minus coordinates of the entire city---the {\em relationship} between the current city and the other cities---are entered into GNN.
For the matrix input TSP, however, a row originally includes information of the relationship.
This means that, in our model, GNN works as same as both GNN and LSTM in GPN alone.
This is why we propose a model which eliminates LSTM from an encoder.
We will verify the possibility of removing LSTM in Sec\ref{sec:ex}.

The output feature vector of the encoder, called the {\em context}, is forwarded to the decoder.
Decoder has an attention layer that returns probabilistic distribution of each cities, and through the softmax layer we obtain the next choice.
Repeating this operation $n$ times, we finally obtain a permutation $\sigma$ of the cities.

\subsection{GPN for QAP}
\label{method-QAP}

\subsubsection{Distance-Flow Product Matrix}
\begin{figure}[t]
\centering
\includegraphics[width =\hsize]{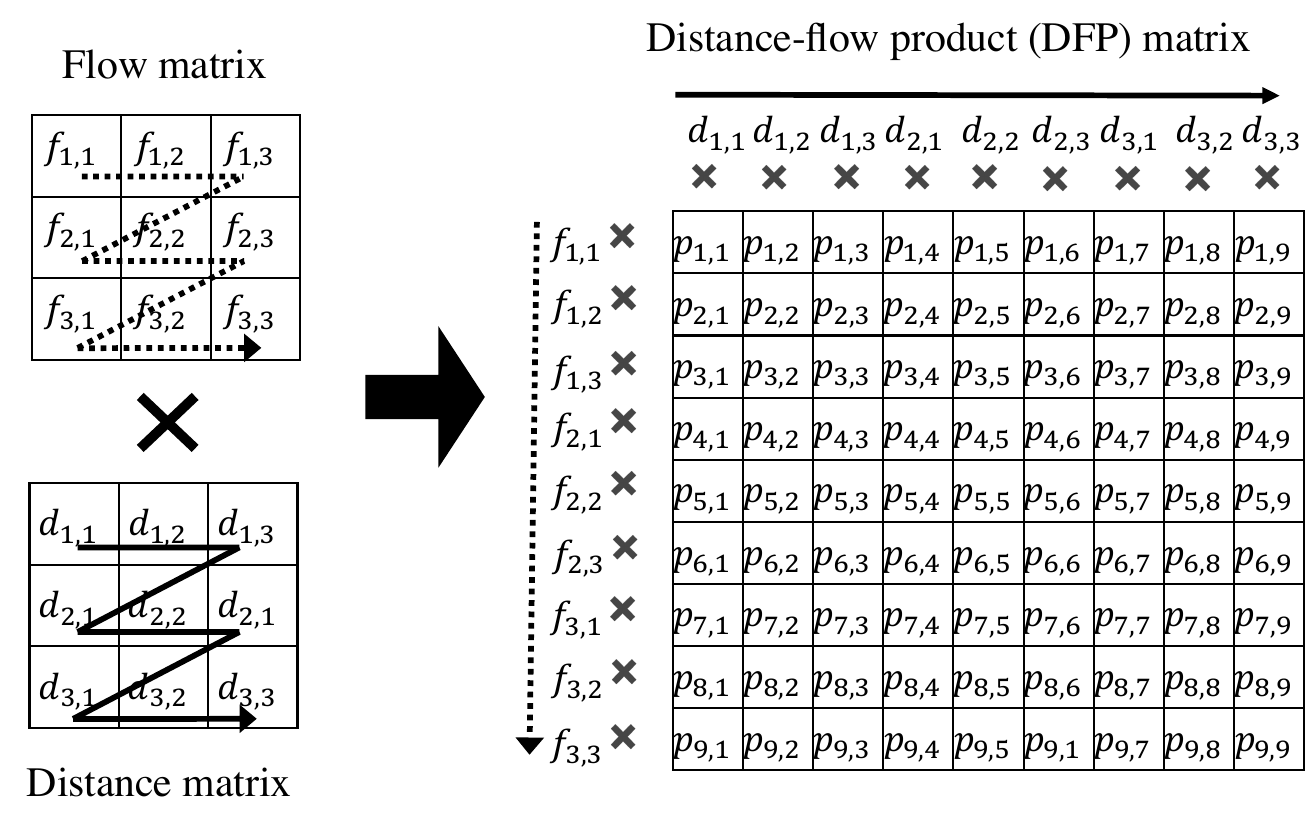}
\caption{Distance-Flow Product (DFP) matrix obtained from distance and flow matrices.}
\label{fig:qap1}
\end{figure}

As shown in Eq. (\ref{eq-qap}), the objective function of QAP is the sum of the distance-flow products. Thus, we introduce the {\em distance-flow product} ({\em DFP}) matrix as an input to GPN.
DFP matrix is obtained as depicted in Fig.~\ref{fig:qap1}.
The row of DFP matrix corresponds to the row-major order of the elements of the distance matrix.
Similarly the column corresponds to the row-major order of the elements of the flow matrix.
Based on this setting, the element becomes the product of the values of the row and the column.
For example, $p_{2,3}=d_{1,2} \cdot f_{1,3}$ in Fig.~\ref{fig:qap1}.

\begin{figure}[t]
\centering
\includegraphics[width =\hsize]{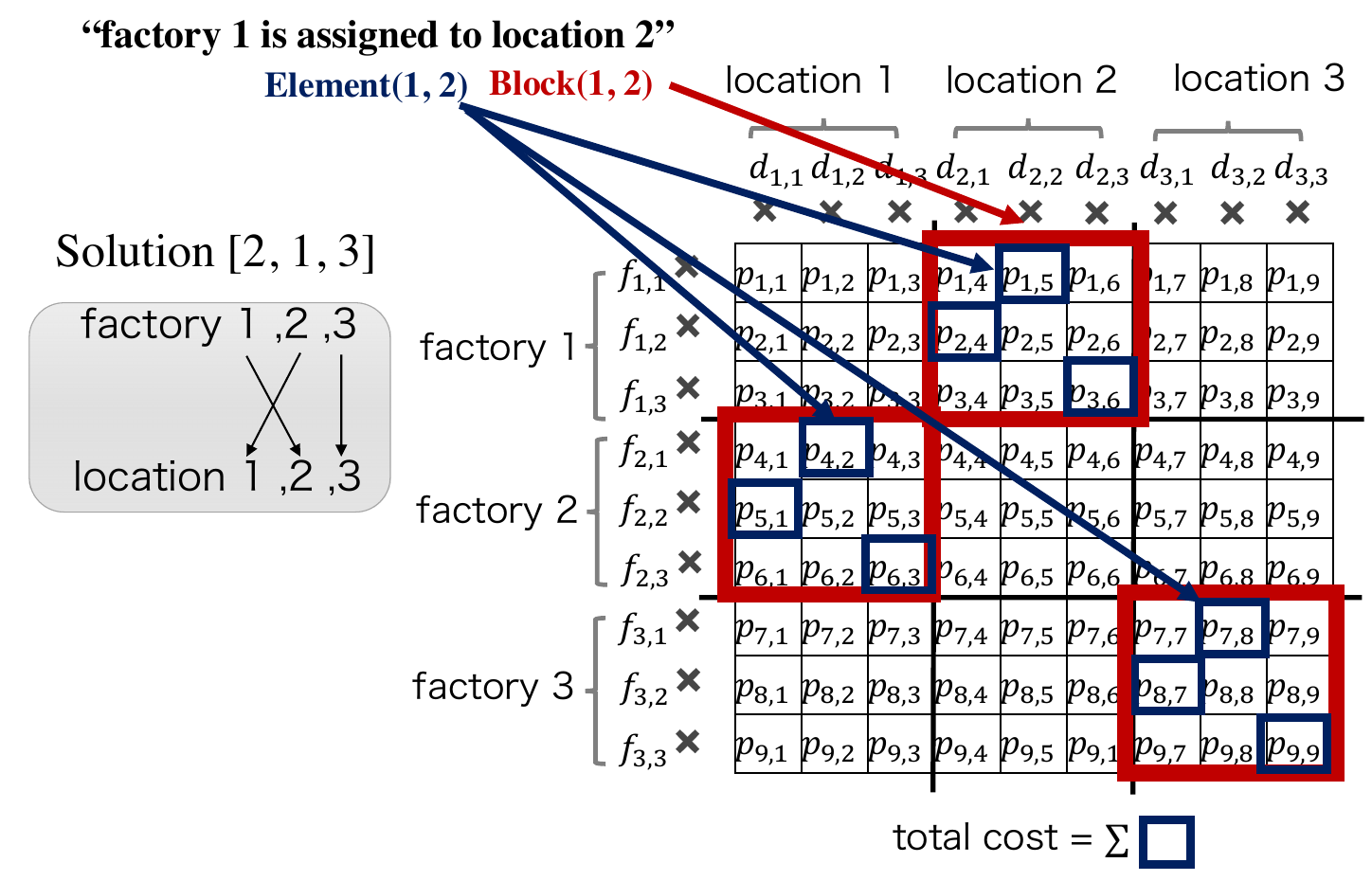}
\caption{The added costs in the DFP matrix when solution is $[2,1,3]$.}
\label{fig:qap_block}
\end{figure}

Suppose that we have a QAP instance $n=3$ and a solution $[2,1,3]$.
This denotes that factories 1, 2, 3 are assigned to locations 2, 1, 3, respectively.
Thus, the total cost is $d_{2,2} f_{1,1} + d_{2,1} f_{1,2} + d_{2,3} f_{1,3} + d_{1,1} f_{2,2} + d_{1,2} f_{2,1} + d_{1,3} f_{2,3} + d_{3,3} f_{3,3} + d_{3,2} f_{3,1} + d_{3,1} f_{3,2}$.
In this case, the total cost is equal to the sum of blue boxes in Fig.~\ref{fig:qap_block}.
We can divide the elements into $n \times n$ {\em blocks} written by the red square in the figure.
The assignment from factory 1 to location 2 identifies the block that includes the added elements as Block(1,2) as shown in the figure. Similarly, the remaining assignments identify them as Block(2,1) and Block(3,3).
We can also divide the elements in each block into $n \times n$ {\em elements} written by the blue square.
In each block, the assignments identify the added elements as Element(2,1), Element(2,1), and Element(3,3).
Obviously, the identified blocks and the identified elements are self-similar.

\subsubsection{Two-stage GPN for QAP}

\begin{figure*}[t]
\centering
\includegraphics[width =\hsize]{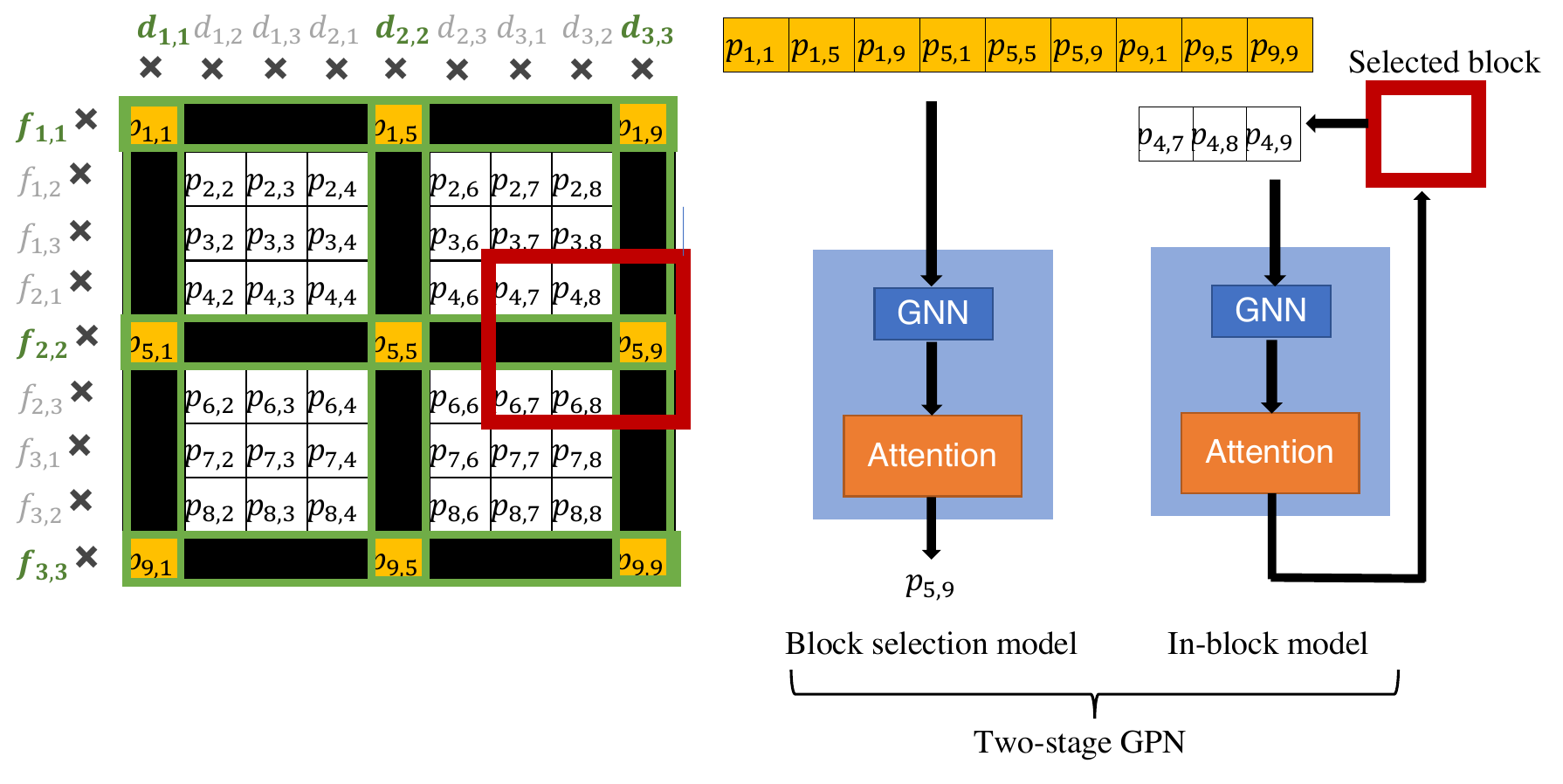}
\caption{Two-stage GPN proposed for solving QAP.}
\label{fig:qap_1}
\end{figure*}

A model for QAP should output the permutation $\pi$ with $n$ elements, while the size of the DFP matrix is $n^2 \times n^2$.
However, GPN generates $n^2$ elements from an $n^2 \times n^2$ matrix.
Thus, we need to change a model so that it generates $n$ elements from an $n^2 \times n^2$ matrix. We propose a {\em two-stage GPN} for this change.

The proposed two-stage GPN is depicted in Fig.~\ref{fig:qap_1}. It consists of two GPN, the {\em block selection} model and the {\em in-block} model. Intuitively, the block selection model selects the focused block, and then the in-block model generates the solution using the elements in the selected block.

\begin{figure}
\centering
\includegraphics[width=\hsize]{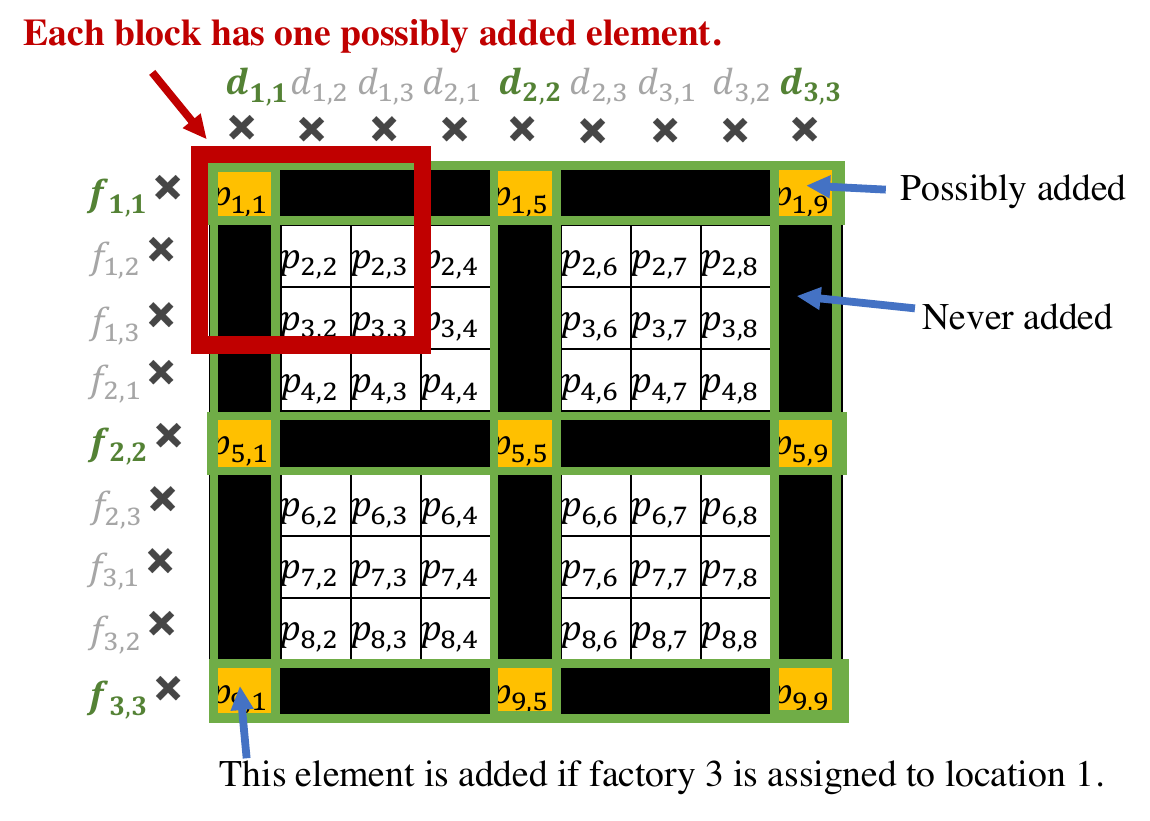}
\caption{Possibly added elements in the lines of $d_{i,i}$ and $f_{i,i}$.}
\label{fig:one_assign}
\end{figure}

Let us focus on the lines of $d_{i,i}$ and $f_{i, i}$ in the DFP matrix, as shown by the green squares in Fig.~\ref{fig:one_assign}. In these lines, the number of possibly added elements to the cost is only $n^2$, while they have $2n^3 - n^2$ elements in total. This is because $f_{i, i}$ and $d_{i, i}$ are respectively multiplied to $d_{j, k}$ and $f_{j, k}$ only if $j = k$ holds. Also as shown in the figure, each block has one possibly added element. In light of this fact, one GPN called the block selection model selects one element from these possibly added elements. This corresponds to select a block.

After the block is selected, the other model called the in-block model is applied to the selected block. We input the values in the selected block into the in-block model repeatedly and obtain the permutation. Finally, the total costs are computed and the parameters are updated.

\subsubsection{Multiple Models for Various Sparsity}
Our method has a drawback that it is not suitable for sparse QAP instances with many zeros in the input matrix. This is because most of the input values to GPN can possibly be zero, and the model produces inaccurate results.
In order to deal with such sparse instances, we introduce a tailored model for sparse matrices, where
only sparse matrices are used as training data.

\section{Experiments}
\label{sec:ex}

For experiments we implement the proposed models for matrix-input TSP and QAP on AMD EPYC 7502P 32core/64thread CPU and NVIDIA RTX A6000 GPU.
System environment and the hyperparameters are summarized in Table~\ref{gra:3}.

\begin{table}[t]
\centering
 \caption{(a) System environment and (b) hyperparameters.}
 \label{gra:3}
 \subfloat[]{
 \begin{tabular}{lr}
 \hline
 CPU & AMD EPYC 7502p 32core/thread 2.5Hz\\
 GPU & NVIDIA RTX A6000\\
 Host Memory & 128GB ECC Registered DDR4-3200\\
 Device Memory & 48GB GDDR6\\
 OS & Ubuntu 20.04 LTS\\
 \hline
 \end{tabular}
 }\\
 \subfloat[]{
  \begin{tabular}{lr}
  \hline
   Epoch & 10\\
   Batch size & 150\\
   Learning step in epoch & 2500\\
Optimizer & Adam\\
 Learning rate & 1e-3\\
   Learning rate decay &0.96\\
   Input size in the training phase & 50 for TSP and 49 for QAP\\
\hline
  \end{tabular}
  }
\end{table}

\subsection{Experiments for matrix input TSP}\label{sect:eval_tsp}
\label{result-tsp}
Firstly, we evaluate GPN for matrix input TSP by solving TSP instances in TSPLIB~\cite{tsplib, tsplib-url}. 
We compare the costs of the obtained solutions and the execution time for the inference. As well as comparing our model with the original GPN, which can support only Euclidean TSP, we compare our model with that with LSTM for demonstrating the impact of eliminating LSTM.

\begin{table*}[t]
\centering
 \caption{Results for matrix input TSP. Instance corresponds to the name of the problem in TSPlib, in which the number indicates the number of cities, i.e., the problem size. The gap indicates the difference between the best-known solution and the obtained solution.}
 \label{gra:3}
  \begin{tabular}{lrrcrrlrrll}
  \hline
  \multirow{2}{*}{Instance} & \multirow{2}{*}{Best-known costs} &\multicolumn{3}{c}{Original GPN}&\multicolumn{3}{c}{Our GPN with LSTM}&\multicolumn{3}{c}{Our proposed GPN (without LSTM)}\\
    \cline{3-11}
    & & Costs & Gap [\%] & Time [s]& Costs & Gap [\%] & Time [s] & Costs & Gap [\%] & Time [s] \\
    \hline
    eil76 & 538 & {\bf 596.8} & 10.93 & 2.96 & 712.0 & {\bf 32.34} (worst) & 2.78 & 712.0 & {\bf 32.34} (worst) & {\bf 2.34}\\
    eil101 & 629 & {\bf 708.7} & 12.67 & 3.01 & 825.2 & 31.20 & 2.82 & 825.2 & 31.20 & {\bf 2.39}\\
    ch130 & 6110 & {\bf 6864} & 12.35 & 3.09 & 7575 & 23.98 & 2.88 & 7575 & 23.98 & {\bf 2.42}\\
    ch150 & 6528 & {\bf 7173} & 9.878 & 3.13 & 8195 & 25.53 & 2.95 & 8195 & 25.53 & {\bf 2.47}\\
    a280 & 2579 & 3275 & 26.99 & 3.39 & {\bf 3148} & 22.07 & 3.22 & {\bf 3148} & 22.07 & {\bf 2.60}\\
    u574 & 369057 & 47752 & 29.39 & 3.95 & {\bf 46881} & 27.03 & 3.84 & {\bf 46881} & 27.03 & {\bf 3.28}\\
    u1432 & 152970 & 197227 & 28.93 & 4.89 & {\bf 188815} & 23.43 & 5.67 & {\bf 188815} & 23.43 & {\bf 4.80}\\
    gr48 & 50467 & \multicolumn{3}{c}{$\infty$ (not supported)} & 6098 & 20.85 & 2.74 & 6098 & 20.85 & {\bf 2.43}\\
    si175 & 21407 & \multicolumn{3}{c}{$\infty$ (not supported)} & 22263 & 3.998 & 2.98 & 22263 & 3.998 & {\bf 2.37}\\
    si535 & 48450 & \multicolumn{3}{c}{$\infty$ (not supported)} & 50144 & 3.496 & 3.73 & 50144 & 3.496 & {\bf 2.47}\\
    si1032 & 92650 & \multicolumn{3}{c}{$\infty$ (not supported)} & 94571 & {\bf 2.073} (best) & 4.80 & 94571 & {\bf 2.073} (best) & {\bf 3.46}\\
    \hline
  \end{tabular}
\end{table*}

The results are shown in Table~\ref{gra:3}.
Note that gr48, si75, si535, and si1032 are not solved by the GPN method because gr48 is represented by a lower triangular matrix and si75, si535, and si1032 are represented by an upper triangular matrix.
The results show that the gap of the solution obtained by our GPN ranges from 2\% to 32\%.
In addition, while the solutions obtained by our GPN and the method before LSTM removal are the same, the execution time of the method without LSTM is faster than that of the method with LSTM.
As described in section \ref{method-TSP}, the LSTM part is not necessary for the proposed method.

The original GPN assumes that the problem size is less than 1000, and the results are more accurate than the proposed method for problems smaller than 1000.
However, when the size of the problem is larger, the accuracy clearly deteriorates.
Therefore, the proposed method has better scalability.

\subsection{Experiments for QAP}

Subsequently, we evaluate our proposed two-stage GPN for QAP by solving QAP instances in QAPlib. 
As with Section~\ref{sect:eval_tsp} we compare the costs of the obtained solutions and the execution time for the inference. The competitors include a greedy algorithm, which selects the minimum of all the parts of the proposed model, and conventional heuristics.

The comparison with a greedy algorithm is shown in Table~\ref{gra:4}. The results show that, in almost all cases, our two-stage GPN provides better solutions than those provided by the greedy algorithm. This suggests that the training is successful. Except for chr instances, the gap between the our obtained solutions and the best-known solutions ranges from 9\% to  30\%. Our model fails to provide approximate solutions for chr instances, because their input matrices include too many zeros. Thus, our proposed method has a limitation that it does not support such instances.

The comparison with conventional heuristic methods, WAITS~\cite{heuristic-1} and SIMD on GPU~\cite{heuristic-2}, is shown in Table~\ref{gra:5}.
WAITS is a method of Whale optimization Algorithm (WA) Integrated with a Tabu Search (WAITS) and is solved by processor Intel Core i5-3317U CPU.
SIMD on GPU denotes a method using a parallel algorithm that employs a 2-opt heuristic and GPU.
The results demonstrate that our two-stage GPN outperforms conventional heuristic methods in terms of the execution time, while the solution quality is inferior to conventional methods. In the case of tai50a, our method is 50.5 times faster than WAITS.

\begin{table*}[t]
\centering
 \caption{Results for QAP.}
 \label{gra:4}
  \begin{tabular}{lrrcccc}
  \hline
  \multirow{2}{*}{Instance} & \multirow{2}{*}{Zero ratio [\%]} & \multirow{2}{*}{Best-known costs} & \multirow{2}{*}{Costs obtained by greedy algorithm} & \multicolumn{3}{c}{Our two-stage GPN} \\
  \cline{5-7}
     & & & & Costs & Gap[\%] & Time [s] \\
\hline
    had12 & 15.97 & 1652 & 1976 & {\bf 1838} & 11.26 & 3.029 \\
    had16 & 12.11 & 3720 & 4358 & {\bf 4104} & 10.32 & 3.035 \\
    had20 & 9.750 & 6922 & 8012 & {\bf 7722} & 11.56 & 3.097 \\
    Bur26a & 22.34 & 5426670 & 6054141 & {\bf 5936776} & 9.40 & 3.050 \\
    Bur26b & 22.34 & 4226496 & 4301616 & {\bf 4109222} & 10.71 & 3.065 \\
    nug12 & 42.70 & 578	& 878 & {\bf 756} & 30.79 & 3.029 \\
    nug30 & 37.06 & 6124 & 8004 & {\bf 7798} & 27.34 & 3.062 \\
    chr12a & 86.20 & 9552 & 34548 & {\bf 30144} & 215.6 (failure) & 3.039 \\
    chr12b & 86.20 & 9742 & 41450 & {\bf 40242} & 313.1 (failure) & 3.035 \\
    chr12c & 86.20 & 11156 & {\bf 24066} & 26542 & 137.92 (failure) & 3.042 \\
\hline
\end{tabular}
\end{table*}

\begin{table*}[t]
\centering
 \caption{Comparison with the latest heuristic methods.}
 \label{gra:5}
  \begin{tabular}{lrrccrccrcc}
  \hline
  \multirow{2}{*}{Instance} & \multirow{2}{*}{Best-known costs} & \multicolumn{3}{c}{Our two-stage GPN} & \multicolumn{3}{c}{WAITS\cite{heuristic-1}} & \multicolumn{3}{c}{SIMD on GPU\cite{heuristic-2}}\\
  \cline{3-11}
    & & Costs  & Time [s] & Gap [\%] & Costs & Time [s] & Gap [\%] & Costs & Time [s] & Gap [\%] \\
\hline
    tai30a & 1818146 & 2160408 & {\bf 3.059} & 18.82 & 1834200.33 & 29.76 & 0.883 & 1838184 & 3.84 & 1.102 \\
    tai40a & 3139370 & 3749304 & {\bf 3.074} & 19.43 & 3168510.8 & 48.88 & 0.928 & 3187882 & 11.83 & 1.545 \\
    tai50a & 4938796 & 5799408 & {\bf 3.134} & 17.43 & 5025226.33 & 204 & 1.750 & 5026692 & 29.40 & 1.780  \\
    tai80a & 13557864 & 15596926 & {\bf 3.229} & 15.04 & 13869196.2 & 1157 & 2.296 & 13833332 & 202.11 & 2.032 \\
    tai100a & 21125314 & 24028820 & {\bf 3.419} & 13.74 & 21600484.25 & 1201 & 2.249 & 21550036 & 501.65 & 2.010 \\
    lipa70a& 169755 & 173577 & {\bf 3.166} & 2.25 & 170396 & 277.32 & 0.378 & 171068 & 117.08 & 0.773 \\
    lipa90a& 360630 & 367381 & {\bf 3.306} & 1.87 & 362554.8 & 725.03 & 0.534 & 362948 & 327.19 &0.643 \\

\hline
\end{tabular}
\end{table*}

\section{Conclusions}\label{sect:conc}
In this paper, we have proposed the two-stage GPN, an extended GPN that approximately solves QAP by reinforcement learning. To this end, we firstly extend an original GPN, which solves only Euclidean TSP, so that it can solve matrix-input TSP. We have demonstrated that the extended GPN can approximately solve matrix-input TSP, and eliminating LSTM accelerates the inference without decreasing the accuracy.
Subsequent extension results in our proposed two-stage GPN.
The experimental results show that our GPN provides approximate solutions better than greedy algorithm, and the execution time is shorter than that of conventional heuristic methods, even with GPU parallelization.
Our two-stage GPN is publicly available~\cite{github}.

\section*{Acknowledgment}
This work was supported by JSPS KAKENHI Grant Numbers JP23H05489 and JP22H05193.

\bibliographystyle{ieeetr}
\bibliography{ref}

\end{document}